\documentclass{article}

\usepackage[preprint]{neurips_2023}

\usepackage[utf8]{inputenc} 
\usepackage[T1]{fontenc}    
\usepackage{hyperref}       
\usepackage{url}            
\usepackage{booktabs}       
\usepackage{amsfonts}       
\usepackage{nicefrac}       
\usepackage{microtype}      
\usepackage{xcolor}         
\usepackage{amsmath}
\usepackage{amssymb}
\usepackage{longtable}
\usepackage{multirow}
\usepackage{algorithm}
\usepackage{algpseudocode}
\usepackage{listings}
\usepackage{tabularx}
\usepackage{lipsum}
\usepackage{booktabs}
\usepackage{graphicx}
\usepackage{listings}
\usepackage{booktabs}
\usepackage{mdframed}
\usepackage{etoolbox}
\mdfsetup{
   frametitlealignment=\center,
   frametitlebackgroundcolor=gray!30
}
\usepackage{caption}
\newenvironment{dialogue}{\begin{description}}{\end{description}}
\newcommand{\DOCTOR}[1]{\item[DOCTOR:] #1}
\newcommand{\PATIENT}[1]{\item[PATIENT:] #1}

\title{Clinical Camel: An Open Expert-Level Medical Language Model with Dialogue-Based Knowledge Encoding}

\author{
  Augustin Toma$^{1,2,*}$ \quad Patrick R. Lawler$^{3,4,5}$ \quad Jimmy Ba$^{1,6}$ \quad Rahul G. Krishnan$^{1,6,7}$\\
  \textbf{Barry B Rubin}$^{3}$ \quad \textbf{Bo Wang}$^{1,3,6,7,8,*,\dagger}$\\
  $^1$Vector Institute for Artificial Intelligence, Toronto, Canada\\
  $^2$Department of Medical Biophysics, University of Toronto, Toronto, Canada\\
  $^3$Peter Munk Cardiac Centre, University Health Network, Toronto, Canada\\
  $^4$McGill University, Montreal, Canada\\
  $^5$Division of Cardiology, University of Toronto, Toronto, Canada\\
  $^6$Department of Computer Science, University of Toronto, Toronto, Canada\\
  $^7$Department of Laboratory Medicine and Pathobiology, University of Toronto, Toronto, Canada\\
  $^8$AI Hub, University Health Network, Toronto, Canada\\
  \texttt{augustin.toma@mail.utoronto.ca} \quad \texttt{bowang@vectorinstitute.ai}
}

\begin{document}

\maketitle
\renewcommand{\thefootnote}{\fnsymbol{footnote}}
\footnotetext[1]{core contributors}
\footnotetext[2]{corresponding author}
\renewcommand{\thefootnote}{\arabic{footnote}}

\
\begin{abstract}
We present Clinical Camel, an open large language model (LLM) explicitly tailored for clinical research. Fine-tuned from LLaMA-2 using QLoRA, Clinical Camel achieves state-of-the-art performance across medical benchmarks among openly available medical LLMs. Leveraging efficient single-GPU training, Clinical Camel surpasses GPT-3.5 in five-shot evaluations on all assessed benchmarks, including 64.3\% on the USMLE Sample Exam (compared to 58.5\% for GPT-3.5), 77.9\% on PubMedQA (compared to 60.2\%), 60.7\% on MedQA (compared to 53.6\%), and 54.2\% on MedMCQA (compared to 51.0\%). In addition to these benchmarks, Clinical Camel demonstrates its broader capabilities, such as synthesizing plausible clinical notes. This work introduces dialogue-based knowledge encoding, a novel method to synthesize conversational data from dense medical texts. While benchmark results are encouraging, extensive and rigorous human evaluation across diverse clinical scenarios is imperative to ascertain safety before implementation. By openly sharing Clinical Camel, we hope to foster transparent and collaborative research, working towards the safe integration of LLMs within the healthcare domain. Significant challenges concerning reliability, bias, and the potential for outdated knowledge persist. Nonetheless, the transparency provided by an open approach reinforces the scientific rigor essential for future clinical applications.\end{abstract}
\clearpage

\section{Introduction} \label{sec:introduction}
 
Large language models (LLMs), such as GPT-4, have demonstrated remarkable capabilities in various applications. However, their deployment in healthcare settings raises concerns due to their proprietary nature, particularly regarding privacy, stability, and transparency. Although open medical LLMs exist, they fall short in performance compared to proprietary alternatives and offer limited context lengths, restricting their use cases.

The performance gap between proprietary and open models is concerning in healthcare, as the latter allows for more rigorous evaluation and validation. Ensuring the safe integration of LLMs into clinical care requires thorough validation, which is not feasible with the current landscape of proprietary models. Moreover, challenges arise when sending healthcare data to private companies, highlighting the value of institutions being able to serve their own models for reliable and safe access. 

We introduce Clinical Camel, an openly available and high-performing medical LLM fine-tuned from LLaMA-2\citep{touvron2023llama} to address these issues. Clinical Camel is trained via QLoRA\citep{dettmers2023qlora} on a single commercial GPU, enabling it to surpass GPT-3.5 in performance on standardized medical benchmarks: biomedical subsections of the MMLU, MedMCQA, MedQA, PubMedQA, and the USMLE sample exam. We introduce a novel method called Dialogue-Based Knowledge Encoding (DBKE) to develop our training corpus, which converts dense clinical review articles into synthetic conversations.

Our research demonstrates the feasibility of efficiently fine-tuning domain-specific LLMs without the need for massive datasets or computing power. Clinical Camel is an example of open medical LLMs that compare favorably with proprietary counterparts. Nonetheless, evaluating LLMs in healthcare remains challenging, and performance on automated benchmarks does not equate to clinical utility or safety. 

By making Clinical Camel openly available for research, we aim to promote further investigation into the safe integration of LLMs into clinical care and contribute to the advancements of machine learning applications in health.

\subsection{The Application of Large Language Models in Healthcare}

LLMs have a broad scope of potential medical applications due to their ability to process unstructured clinical text; these range from automated clinical note creation and patient record summarization to more advanced tasks like clinical decision support, medical triaging, patient counseling, and medical education. These applications could improve healthcare delivery and access for providers and patients if proven effective.

Proprietary models like OpenAI's GPT-3.5 and GPT-4 demonstrate strong performance on medical benchmarks without domain-specific fine-tuning\cite{nori_capabilities_2023}. GPT-4's capabilities have prompted efforts to integrate it into clinical care, but sending healthcare data to private servers creates access equity issues globally. Critically, rigorously studying proprietary models is challenging. For example, OpenAI updates models on a three-month basis, complicating deployment in patient care where even small prompt changes can drastically alter outputs.

Google's Med-PaLM 2 surpassed GPT-4  when tested with an ensemble refinement strategy (an inference-heavy prompting strategy requiring 44 generations), demonstrating superior performance on MedQA, PubMedQA, MMLU-Professional Medicine, and MMLU-College Medicine benchmarks\cite{singhal2023expertlevel}. Human evaluations also showed physicians and laypeople preferred Med-PaLM 2 answers over physician-generated responses- although the human evaluation group was modestly sized with 15 physicians and six laypersons. The Med-PaLM 2 work is commendable for going beyond automated benchmarks; however, Med-PaLM-2 remains unavailable publicly, preventing external validation of these results. 

The inability to rigorously study proprietary models due to the lack of public information, access, and privacy constraints motivates the development of open  alternatives. High-performing publicly available models will enhance access and enable the rigorous evaluation needed for the safe clinical integration of LLMs.

\section{Open Medical Language Models: Pushing for Transparency and Better Public Health Outcomes}

Several open medical language models have been released, including  MedAlpaca\citep{han_medalpaca_2023} and ChatDoctor\citep{li_chatdoctor_2023}. Limited benchmark evaluations for these models have been made, and no rigorous comparisons have been made to proprietary models such as GPT-3.5/4.

ChatDoctor was fine-tuned on online physician-patient dialogues and compared favorably to GPT-3.5 on BERTScore metrics - which were calculated by comparing the BERTScore of ChatDoctors responses on a dataset comprising of patient questions and answers; however, no other benchmarks were evaluated. Its short context length of 512 tokens restricts utility beyond question-answering.

MedAlpaca reported high performance on the USMLE self-assessment test. However, it also has a trained context length of 512 tokens. A parameter-efficient variant was trained alongside a fully fine-tuned version; however, it significantly underperformed. No other benchmark results were reported. 

In conclusion, while existing open models show promise, benchmark evaluations have been limited and lack comparisons to proprietary models. Their short contexts likely restrict utility as well. In contrast, Clinical Camel has an expanded 4096 token context length and can perform tasks beyond question answering. Consequently, Clinical Camel represents a substantial advancement for deploying large language models in healthcare.

\section{Methodology} \label{sec:methodology}

\subsection{Dialogue-Based Knowledge Encoding}

Our work introduces Dialogue-Based Knowledge Encoding (DBKE), a method designed to transform input text into a multi-turn dialogue. The methodology we have developed acts as a form of domain adaptation that we hypothesize strengthens the recall capabilities of the downstream conversational models. DBKE allows us to convert dense medical literature into dialogues and instill soft alignment. 
 
The DBKE process consists of dialogue creation and student model training. The process is initiated with a dense knowledge text input, paired with an input prompt containing alignment constraints and instructions for generating a dialogue. A teacher model, denoted by $M_T$, generates a dialogue based on the provided context while following the constraints stated in the prompt. The generated dialogue is then used as a transformed training text for fine-tuning a student model, denoted by $M_S$. 

We illustrate the steps of the DBKE methodology in Algorithm \ref{dbke}:

\begin{algorithm}
\caption{Dialogue-Based Knowledge Encoding (DBKE)}\label{dbke}
\begin{algorithmic}[1]
\Procedure{DBKE}{$T, P, M_T, M_S$} \Comment{T is input text, P is prompt (containing alignment rules), $M_T$ is teacher model, $M_S$ is student model}
\For{each target text $t_i$ in $T$}
    \State $D \gets$ Generate a dialogue from $t_i$ using $M_T$ and $P$ \Comment{Teacher model generates multi-turn dialogue}
\EndFor
\State Fine-tune $M_S$ on $D$, masking user's inputs during training
\State \Return $M_S$ \Comment{Return the fine-tuned student model}
\EndProcedure
\end{algorithmic}
\end{algorithm}

\begin{figure}[h]
\centering
\includegraphics[width=0.8\textwidth]{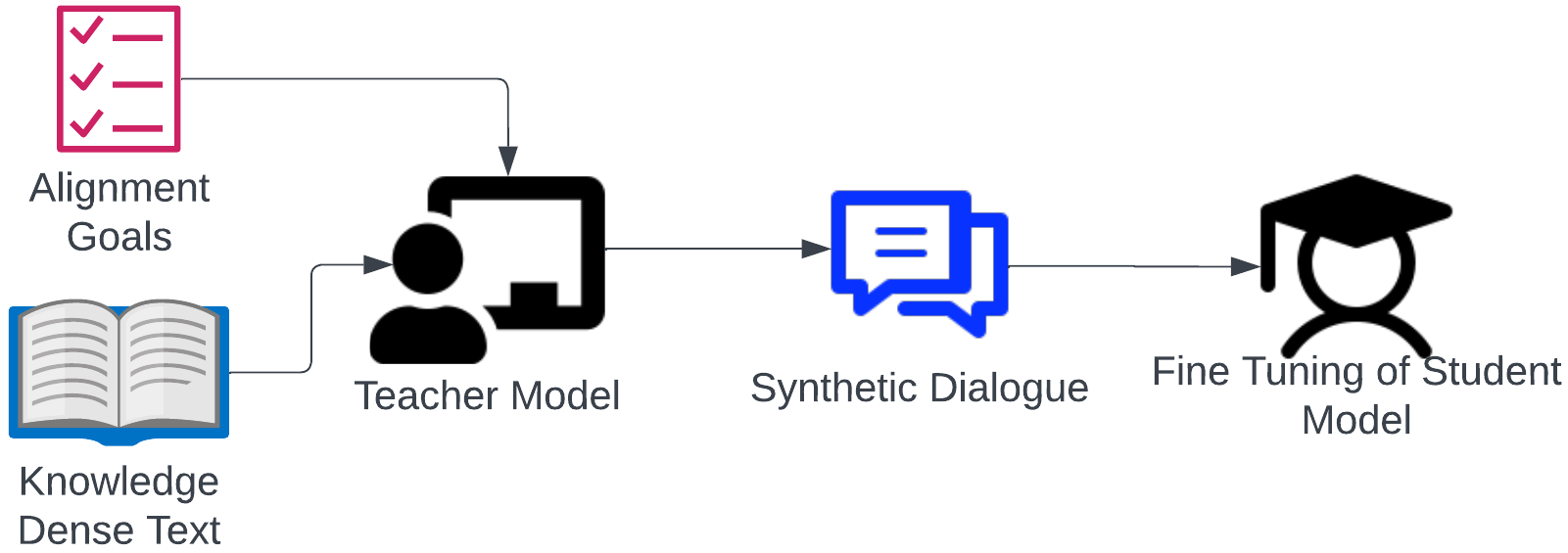}
\caption{Schematic representation of the Dialogue-Based Knowledge Encoding (DBKE) methodology. The process starts with a knowledge-dense input text $T$ and a prompt $P$ containing alignment constraints. The teacher model $M_T$ then generates a multi-turn dialogue $D$, which is used to fine-tune the student model $M_S$. The result is a fine-tuned student model capable of improved conversational performance.}
\label{fig:dbke_fig}
\end{figure}

The DBKE method combines knowledge encoding with soft behavioral alignment. Although not strictly enforced, the alignment constraints embedded in the input prompt guide the generated output of medical models. For example – these constraints could instruct the model to gather more information before suggesting diagnoses. The alignment objectives of these models can be modified to cater to the requirements of specific domains. See \ref{app:example_dbke} for an example. 

\subsection{Dataset} \label{sec:dataset}

We use data from the ShareGPT project\citep{noauthor_sharegpt_nodate}, data from the MedQA training set\citep{jin_what_2020}, and clinical review articles that are available in the public domain from PubMed published before 2021 to minimize the over-representation of COVID-19 related content. The clinical review articles are transformed through the DBKE process to produce synthetic dialogues. The dataset is truncated to 4096 tokens, and non-English text is filtered out. 

\begin{table}[h]
\caption{Summary of datasets in Clinical Camel}
\centering
\begin{tabularx}{\textwidth}{lXX}
\toprule
\textbf{Namae} & \textbf{Description} & \textbf{Preprocessing} \\
\midrule
ShareGPT & Multi-step conversation & Removed non-English text, segmented conversations (4096 tokens), filtered degenerate conversations \\
\addlinespace
Clinical Articles & 20,000 pre-2021 open-access articles & Transformed into 100,000 dialogues (5 utterance exchanges avg.) \\
\addlinespace
MedQA & 4000 randomly selected (10,178 multiple-choice questions pool) & Transformed into dialogue by retrieving relevant source articles and prompting GPT-4 to produce detailed justification for the correct answer from retrieved articles \\
\bottomrule
\end{tabularx}
\label{tab:datasets}
\end{table}

Table \ref{tab:datasets} provides an overview of the datasets used in developing the Clinical Camel model, including their description and preprocessing steps. The ShareGPT data includes general multi-step conversations and comprises 70,000 conversations before preprocessing. Clinical articles include 20,000 open-access clinical articles from various sources published before 2021 that were transformed into 100,000 multi-step dialogues. The MedQA training set has 10,178 multiple-choice questions with non-descriptive answers. We processed a subset of 4000 into dialogues using retrieval augmented generation to identify relevant source texts and provide the correct answer to guide the model's response. The model is encouraged to explain why a particular option is correct and why other options are wrong.

\subsection{Clinical Camel} \label{sec:clinicalcamel}

The LLaMA-2 models serve as the foundation for developing Clinical Camel. We trained 13B and 70B parameter variants using the same dataset. The training utilized QLoRA with masking of human input. This approach enabled training Clinical Camel on a single H100 GPU. Training was conducted for one epoch with the parameters specified in Table~\ref{tab:training_parameters}

\begin{table}[h]
\centering
\caption{Training Parameters}
\label{tab:training_parameters}
\begin{tabular}{lll}
\toprule
\textbf{Parameter} & \textbf{13B Model} & \textbf{70B Model} \\
\midrule
Sequence Length & 4096 & 4096 \\
Lora\_r & 64 & 64 \\
Lora\_alpha & 16 & 16 \\
Lora\_dropout & 0.00 & 0.00 \\
Lora\_target\_modules & All linear layers & All linear layers \\
Gradient Accumulation Steps & 16 & 32 \\
Mini-batch Size & 1 & 1 \\
Number of Epochs & 1 & 1 \\
Optimizer & paged\_adamw\_32bit & paged\_adamw\_32bit \\
Learning Rate Scheduler & Cosine & Cosine \\
Learning Rate & 0.0002 & 0.0001 \\
\bottomrule
\end{tabular}
\end{table}

\section{Evaluation} \label{sec:evaluation}

We evaluated Clinical Camel's performance on standard medical benchmarks in zero- and five-shot settings. Table \ref{tab:zero_shot} presents the zero-shot results compared to GPT-3.5 and GPT-4. Table \ref{tab:five_shot} shows the five-shot results alongside GPT-3.5, GPT-4, and Med-PaLM 2.

The GPT and Med-PaLM-2 scores were sourced from studies by Microsoft\citep{nori_capabilities_2023} and Google\citep{singhal2023expertlevel}. Clinical Camel scores were computed using EleutherAI's evaluation framework\citep{eval-harness}, which compares response likelihoods, we report the accuracy scores for all benchmarks.

In five-shot testing, our model outperforms GPT-3.5 across all metrics. However, it currently falls short of GPT-4 and Med-PaLM 2, except surpassing GPT-4 on PubMedQA.

\begin{table}[h]
\centering
\caption{Performance of Clinical Camel-13B (C13), Clinical Camel-70B (C70), GPT3.5, and GPT4 on various medical datasets in a zero-shot setting.}
\label{tab:zero_shot}
\begin{tabular}{lcccc}
\toprule
\textbf{Dataset} & \textbf{C13 (0-shot)} & \textbf{C70 (0-shot)} & \textbf{GPT3.5 (0-shot)} & \textbf{GPT4 (0-shot)} \\
\midrule
MMLU Anatomy & 50.4 & 62.2 & 56.3 & 80.0 \\
MMLU Clinical Knowledge & 54.0 & 69.8 & 69.8 & 86.0 \\
MMLU College Biology & 54.9 & 79.2 & 72.2 & 95.1 \\
MMLU College Medicine & 48.0 & 67.0 & 61.3 & 76.9 \\
MMLU Medical Genetics & 59.0 & 69.0 & 70.0 & 91.0 \\
MMLU Professional Medicine & 51.8 & 71.3 & 70.2 & 93.0 \\
MedMCQA & 39.1 & 47.0 & 50.1 & 69.5 \\
MedQA (USMLE) & 34.4 & 53.4 & 50.8 & 78.9 \\
PubMedQA & 72.9 & 74.3 & 71.6 & 75.2 \\
USMLE Sample Exam & 26.9 & 54.3 & 49.2 & 83.2 \\
\bottomrule
\end{tabular}
\end{table}

\begin{table}[h]
\centering
\caption{Performance of Clinical Camel-13B (C13), Clinical Camel-70B (C70), GPT3.5, GPT4, and Med-PaLM 2 on various medical datasets in a five-shot setting.}
\label{tab:five_shot}
\small
\begin{tabular}{>{\raggedright\arraybackslash}p{2.5cm}cccccc}
\toprule
\textbf{Dataset} & \textbf{C13} & \textbf{C70} & \textbf{GPT3.5} & \textbf{GPT4} & \textbf{Med-PaLM 2} \\
 & \textbf{(5-shot)} & \textbf{(5-shot)} & \textbf{(5-shot)} & \textbf{(5-shot)} & \textbf{(5-shot)} \\
\midrule
MMLU Anatomy & 48.2 & 65.2 & 60.7 & 80.0 & 77.8 \\
MMLU Clinical Knowledge & 60.4 & 72.8 & 68.7 & 86.4 & 88.3 \\
MMLU College Biology & 59.0 & 81.2 & 72.9 & 93.8 & 94.4 \\
MMLU College Medicine & 52.6 & 68.2 & 63.6 & 76.3 & 80.9 \\
MMLU Medical Genetics & 59.0 & 69.0 & 68.0 & 92.0 & 90.0 \\
MMLU Professional Medicine & 53.3 & 75.0 & 69.8 & 93.8 & 95.2 \\
MedMCQA & 44.8 & 54.2 & 51.0 & 72.4 & 71.3 \\
MedQA (USMLE) & 45.2 & 60.7 & 53.6 & 81.4 & 79.7 \\
PubMedQA & 74.8 & 77.9 & 60.2 & 74.4 & 79.2 \\
USMLE Sample Exam & 39.5 & 64.3 & 58.5 & 86.6 & - \\
\bottomrule
\end{tabular}
\end{table}
\pagebreak

\section{Capabilities, challenges, and future directions of the Clinical Camel} \label{sec:futurework}

In addition to strong performance on medical question-answering benchmarks, Clinical Camel shows promise for other healthcare applications like automated clinical note generation. As demonstrated in Figure~\ref{t}, the model can effectively synthesize plausible clinical notes from long patient-physician conversations(see Appendix~\ref{app:dialogue}) while adhering to alignment objectives. This ability to handle extended contexts is a crucial capability arising from Clinical Camel's 4096 token limit.

However, several challenges remain in applying Clinical Camel more broadly in healthcare settings. A primary concern is the potential for generating misleading or inappropriate content \citep{ji_survey_2023}. Evaluating model outputs and developing techniques to improve reliability and alignment will be critical for future research directions.

Another challenge stems from updating medical LLMs as knowledge evolves continually. Retraining models on new data requires significant computational resources. Alternative approaches like memory editing \citep{meng_mass-editing_2022} and retrieval-augmented generation \citep{shuster_retrieval_2021} may enable more efficient knowledge updating and will be essential to explore.

Additionally, Clinical Camel is not multi-modal, which is a significant limitation in healthcare. Extending the model to multi-modal inputs could improve its utility for diagnostic and other visual tasks.

We also note that we have yet to systematically evaluate the effectiveness of DBKE compared to other methods of processing training data. Therefore we cannot make definitive statements about the effectiveness of DBKE. 

In summary, while Clinical Camel demonstrates promising capabilities on medical benchmarks, further research is needed to improve reliability, update knowledge, and incorporate multi-modal data. As an open model, Clinical Camel will facilitate this continued study into safely and effectively applying LLMs in healthcare.
\pagebreak

\begin{mdframed}[frametitle={Transcribed Note}]
\begin{description}
    \item[ID:] name, age, and gender not mentioned
    \item[REASON FOR VISIT:] Leg pain and difficulty walking
    \item[PAST MEDICAL HISTORY:] Borderline diabetes for 20 years
    \item[HOME MEDICATIONS:] Diabetes medication (not mentioned)
    \item[ALLERGIES:] Shellfish
    \item[FAMILY HISTORY:]  Multiple family members died in their sleep
    \item[SOCIAL HISTORY:]
    \begin{itemize}
        \item Smoking: 1-2 packs per day
        \item Alcohol: Not mentioned
        \item Occupation: Not mentioned
    \end{itemize}
    \item[HISTORY OF PRESENT ILLNESS:] The patient has been experiencing difficulty walking for 20 years, with pain in the left leg, starting in the ankle and calf, and sometimes affecting both legs. The pain is relieved by rest. The patient also experiences numbness in the toes and fingers, and sometimes wakes up at night due to numbness and tingling sensations. The patient denies chest pain but reports heartburn sensations after walking. The patient has shortness of breath but it is not the primary reason for stopping walking.
    \item[PHYSICAL EXAM:] Not mentioned
    \item[RESULTS:] Ultrasound of legs: Significant narrowing in the main artery of the left leg
    \item[ASSESSMENT AND PLAN:]
    \begin{itemize}
        \item Schedule a CAT scan of leg arteries to determine the extent of blockages or narrowings
        \item Schedule an ultrasound of the heart and a stress test (dopamine echocardiogram) to assess heart function
        \item Discuss treatment options based on test results, including potential interventions to improve circulation in the leg and prevent further complications
        \item Encourage the patient to bring family members to the next appointment for support and discussion of treatment options
    \end{itemize}
\end{description}
\end{mdframed}
\captionof{figure}{Clinical note generated by Clinical Camel from the dialogue in Appendix \ref{app:dialogue}}
\label{t}

\subsection{Bridging the Divide}
Recent advances in parameter-efficient training methods, along with the release of models like MetaAI's LLaMA, have led to rapid improvements in open language models; as a result, Clinical Camel outperforms GPT-3.5 on medical benchmarks, despite being trained on a single commercial GPU. However, a significant gap remains compared to top-performing models such as GPT-4 and Med-PaLM-2. 

Open initiatives have the potential to continue closing this gap through data rather than computing. In many countries, public health institutions control massive datasets that could help train open medical models. Collaborations between public and private entities can enable responsible access to these records, creating fine-tuned models based on anonymized electronic health record data - this stands in contrast to the fragmented efforts undertaken by competing private companies.

The strategic harnessing of public healthcare data resources could help democratize model development for equitable public benefit. With patient consent and privacy techniques, health records could be used to co-develop open models designed for patients first.

Additionally, open development enables transparency and collaboration fundamental for scientific study. Openness facilitates engaging diverse experts and patients to provide critical input. 

In summary, while open model development efforts may lack the computing scale of private corporations, they could leverage extensive public data. Responsible data initiatives could help democratize development toward open models finely tuned for serving all patients.

\section{Ethical Considerations}

Deploying LLMs like Clinical Camel raises many ethical concerns \citep{harrer_attention_2023}; paramount is patient safety, as these models can generate misleading or incorrect information, potentially causing inappropriate diagnoses or treatments. Thorough evaluation and real-world testing are essential to ensure safe deployment.

Bias in model outputs, fueled by skewed training data, may lead to unfair outcomes for diverse populations. Proactively assessing and mitigating dataset and output biases is crucial. Any ethical efforts require clear accountability, regular accuracy checks, and comprehensive monitoring and reporting. Deploying healthcare LLMs demands rigorous ethical precautions. Foremost is ensuring patient safety, as inaccurate model outputs risk inappropriate diagnoses or treatments. Extensive evaluation across diverse clinical contexts is essential pre-deployment and ongoing real-world monitoring post-deployment to enable early error detection and prevent patient harm.

Imbalanced training data may fuel model biases yielding unfair outcomes for underrepresented groups. Proactive bias detection and mitigation in datasets and outputs are imperative, alongside mandated ongoing accountability through accuracy benchmarking and progress reporting.

Furthermore, upholding safety and equity requires close collaboration with patients, clinicians, ethicists, and experts from marginalized communities throughout development, centering patient voices.

Clinical Camel is not ready for actual clinical application. By openly releasing the model, we aim to promote the rigorous study needed to integrate similar LLMs safely. Much work remains to evaluate and improve performance across diverse populations and prevent potential harm before clinical use. Transparent development and evaluation of open models like Clinical Camel are essential to realizing benefits while acting in a principled manner.

\section{Conclusion} \label{sec:conclusion}

Clinical Camel demonstrates competitive performance to proprietary LLMs via efficient training, achieving state-of-the-art results among open medical models and surpassing GPT-3.5 on QA benchmarks. However, benchmark metrics alone insufficiently evidence real-world efficacy and safety. Extensive human assessment across diverse clinical contexts is essential pre-deployment, and ongoing monitoring post-deployment, to enable early error detection. Sustained accountability around updating, transparency, and integrating patient perspectives is vital to uphold ethics as applications progress toward practice. By openly releasing Clinical Camel, we aim to promote collaboration on rigorously evaluating LLMs pre-clinically to harness their possibilities for patients safely. However, significant work remains to prevent potential harm before clinical integration. Open development and assessment of models like Clinical Camel is essential to realizing benefits while upholding scientific ethics.

\section{Model Access }

The model can be found online:

Hugging Face: \url{https://huggingface.co/wanglab}

\textbf{Disclaimer}: Please note that users must agree to not use the model for actual patient care, the model is released for research purposes.

\newpage

\newpage
\bibliographystyle{plainnat}
\bibliography{ClinicalCamel}
\pagebreak
\appendix
\section{Appendix A: Dialogue}
\label{app:dialogue}
\begin{dialogue}
\DOCTOR{ How can I help you? }
\PATIENT{ Hi, Dr. You know, it's been 20 years that I've been having this problem. Sorry. But it's really been, it's just been affecting my life. I'm having problems with my toes, I can't feel them. And now I'm having difficulty to walk.}
\DOCTOR{ How far can you walk? }
\PATIENT{ Half a block. }
\DOCTOR{ What happens then? }
\PATIENT{ I start to get some pain in the left leg. I have to stop. And then if I push myself and I have to go because I have somewhere to get, then I can get pain in both legs. Where in your left leg does it start?}
\DOCTOR{ Kind of starts in the ankle and the calf and kind of migrates upwards. If you're sitting around not doing anything, do you ever get pain?}
\PATIENT{ No, I get numbness. }
\DOCTOR{ What is the numbness?}
\PATIENT{ It's in the feet. I can't feel my toes most of the time. }
\DOCTOR{ Does it go up past your ankle, your calf, your knee? }
\PATIENT{ I don't know actually, I've really just noticed it in my toes.}
\DOCTOR{ Ever wake you up at night?}
\PATIENT{ Yes. }
\DOCTOR{ What wakes you up at night? }
\PATIENT{ This weird sensation of numbness and tingling that can sometimes be painful.}
\DOCTOR{ Is it both feet or just one?}
\PATIENT{ Starts off in one, but sometimes it's mostly the left, but sometimes it can be both.}
\DOCTOR{ Do you have diabetes? }
\PATIENT{ Well, kind of. I've been told I've been borderline diabetic for 20 years.}
\DOCTOR{ Do you take medication for diabetes?}
\PATIENT{ I'm supposed to.}
\DOCTOR{ Do you smoke?}
\PATIENT{ I do. }
\DOCTOR{ How much do you smoke? }
\PATIENT{ About a pack or two. High blood pressure? }
\DOCTOR{ High cholesterol? }
\PATIENT{ I don't know, I haven't really seen my family doctor in about five years. I went to the walk-in because of the feet and that's how I ended up here.}
\DOCTOR{ Allergies?}
\PATIENT{ Shellfish.}
\DOCTOR{ Do you have any brothers or sisters? }
\PATIENT{ I do. }
\DOCTOR{ Any of your brothers or sisters or parents have heart problems?}
\PATIENT{ They've all died in their sleep. They've all died in their sleep. }
\DOCTOR{ Do you know, was there ever a post-mortem exam to understand what happened to them?}
\PATIENT{ No. }
\DOCTOR{ Yeah, that's very sad. I'm sorry to hear that. }
\PATIENT{ Thank you. }
\DOCTOR{ Okay. And do you ever get any chest pain?}
\PATIENT{ I get this weird heartburn sensation. }
\DOCTOR{ Tell me about that.}

\PATIENT{ So, if I go for a walk and I have the burning in my feet or the pain and the burning, sometimes feel burning in the stomach. That goes away when I have to rest. }
\DOCTOR{ Do you get sweaty when that happens? }
\PATIENT{ Maybe. Not consistently, but yeah.}
\DOCTOR{ Do you get any pain in one arm or another?}
\PATIENT{ No.}
\DOCTOR{ Does the feeling you have in your stomach go up into your neck or into your head? }
\PATIENT{ No, it's kind of stuck there. It's sort of this burning sensation.}
\DOCTOR{ You're short of breath? }
\PATIENT{ All the time. }
\DOCTOR{ When you walk, what's more likely to stop you from walking? The pain in your left leg or the shortness of breath.}
\PATIENT{ The pain. The pain comes first. I don't notice really the breathing. It's more the pain and then, because I'm sitting quietly, then I notice that I have some heartburn.}
\DOCTOR{ Okay. And have you ever had an episode where you suddenly lost vision in one eye or the other, like a curtain came over your eye?}
\PATIENT{ No.}
\DOCTOR{ Do you ever have any difficulty speaking?}
\PATIENT{ No. }
\DOCTOR{ Any problems moving one arm or one leg?}
\PATIENT{ No.}
\DOCTOR{ Any numbness other than the numbness of your feet?}
\PATIENT{ I have some numbness in my fingers.}
\DOCTOR{ Okay. So we did an ultrasound of your legs and we can see that there's quite a significant narrowing in the main artery in your left leg. Why? So it's maybe because of previous smoking. It may be because of borderline diabetes. It may be that this runs in your family, but it doesn't really matter the why. It's there and we need to do more tests to understand how to treat this because with what you're describing, you just have enough blood flow to keep your leg alive and if we don't improve that, you could end up losing a leg.}
\PATIENT{ So you're telling me I'm going to lose my leg?}
\DOCTOR{ I'm telling you that we have to do some tests so that we can see exactly what's going on and then see if there's a way to improve the circulation in your leg so you don't end up losing a leg. I'm not sure what's going to happen yet. }
\PATIENT{ So is what's happening in my leg also happening in my chest?}
\DOCTOR{ So it could be and we're going to also investigate that. So I'm going to get a CAT scan of your arteries in your legs and that's going to tell me where the blockages or narrowings are.}
\PATIENT{ And what if I don't want anything done? }
\DOCTOR{ So that's fine. It's always the patient's choice about what to do. The way that this works is I give you options and then you tell me what you want to do and as long as I'm satisfied that you really understand what I've told you, then it's completely your choice. It would be helpful if you come in if there's any family members for our next meeting. We can discuss this with other people. And you don't have to decide right this moment, but we do have to decide fairly soon because this can progress. So I'm going to get a CAT scan of your leg arteries. I'm going to get an ultrasound of your heart and a stress test of your heart. Because you have problems walking, I'm going to get a specific type of stress test called a dopamine echocardiogram that you don't have to do any walking. We'll just be able to put this all together and we'll see, do you have any narrowings in your heart arteries? Do you have narrowings in your leg arteries? And then I'll make a recommendation about what to do about this. }
\PATIENT{ Okay, that sounds reasonable.}
\DOCTOR{ Do you have any questions for me?}
\PATIENT{ Do you think that this could be why my siblings all died?}
\DOCTOR{ Yes. It could be that they had narrowed heart arteries. But there's lots of other reasons, so it would really be speculation. What's important is to figure out what's going on with you. }
\end{dialogue}

\section{Appendix B: Example of DBKE}\label{app:example_dbke}
\begin{tabularx}{\textwidth}{|X|X|}
\hline
\multicolumn{2}{|p{\textwidth}|}{\textbf{Prompt}} \\
\multicolumn{2}{|p{\textwidth}|}{\small Create a realistic chat dialogue between a patient and a medical chat bot using the passage provided below 
1. Bot empathetically communicates medical information in a simple manner.
2. Bot admits limitations if unsure about information.
3. Patient inquiries cover diverse topics (test results, medications, physical findings, symptoms) related to the passage.
4. Bot asks follow-up questions for better understanding.
5. Focus is on guiding the patient towards understanding their diagnosis.
6. Bot explains its reasoning upon request.
7. Patient provides lab values, imaging descriptions, or ECG findings explicitly.
8. Bot inquires about patient's medical history, medications, symptoms, lab results, and imaging or ECG findings using non-expert language.
9. Bot explains imaging or ECG features suggestive of a diagnosis without claiming to view images.
10. Bot encourages the patient to consult a healthcare provider for further evaluation, not booking appointments or ordering tests directly.
}\\
\hline
\textbf{Input Text} & \textbf{Output Dialogue} \\
There is significant overlap between the lifestyle and activity recommendations for preventing myocardial infarction and those used for secondary prevention after an initial myocardial infarction...
For primary prevention, lifestyle changes such as engaging in regular physical activity, maintaining a healthy weight, drinking alcohol within recommended limits, and quitting smoking can reduce the risk of cardiovascular disease...
Public health measures such as reducing unhealthy diets through food labeling and marketing requirements, and stimulating physical activity can also help reduce the risk of myocardial infarction at a population level...
Medications such as statins, which lower blood cholesterol, and aspirin have been studied extensively for their efficacy in preventing myocardial infarction... &
Patient: Hi, I just read an article about preventing myocardial infarction... Bot: Of course, I'll do my best to help... Patient: How effective is the influenza vaccine in protecting against myocardial infarction? Bot: Studies have shown that the influenza vaccine can reduce the risk of myocardial infarction by 15-45\%... Patient: What lifestyle changes can I make to reduce my risk of myocardial infarction? Bot: Maintaining a healthy weight, engaging in regular physical activity, quitting smoking, and drinking alcohol within recommended limits are all lifestyle changes... Patient: What kind of diet is recommended? Bot: A healthy diet should include wholegrain starch, fruits, vegetables, fish, unsalted nuts, seeds, and legumes... \\
\hline
\end{tabularx}

\end{document}